\documentclass[runningheads]{llncs}

\usepackage[T1]{fontenc}
\usepackage{graphicx}
\usepackage{cite}

\begin{document}
\title{Cognitive Architecture for Co-Evolutionary Hybrid Intelligence\thanks{This work was carried out under the "Development program of ETU 'LETI' within the framework of the program of strategic academic leadership ''Priority-2030'' No~075-15-2021-1318 on 29 Sept 2021.}
}

\author{Kirill Krinkin\orcidID{0000-0001-5949-7830} \and
Yulia Shichkina\orcidID{0000-0001-7140-1686}}
\authorrunning{K. Krinkin, Y. Shichkina}
%
\institute{Saint-Petersburg Electrotechnical University 'LETI', Saint-Petersburg, Prof.~Popova~5 197022, Russia \\ \email{kirill@krinkin.com, shichkina@etu.ai}}
\maketitle              
\begin{abstract}
This paper\footnote{Presented on AGI-2022 http://agi-conf.org/2022/} questions the feasibility of a strong (general) data-centric artificial intelligence (AI). The disadvantages of this type of intelligence are discussed. As an alternative, the concept of co-evolutionary hybrid intelligence is proposed. It is based on the cognitive interoperability of man and machine. An analysis of existing approaches to the construction of cognitive architectures is given. An architecture that seamlessly incorporates a human into the loop of intelligent problem solving is considered. The article is organized as follows. The first part contains a critique of data-centric intelligent systems. The reasons why it is impossible to create a strong artificial intelligence based on this type of intelligence are indicated. The second part briefly presents the concept of co-evolutionary hybrid intelligence and shows its advantages. The third part gives an overview and analysis of existing cognitive architectures. It is concluded that many of them do not consider humans as part of the intelligent data processing process. The next part discusses the cognitive architecture for co-evolutionary hybrid intelligence, providing integration with humans. It finishes with general conclusions about the feasibility of developing intelligent systems with humans in the problem solving loop.

\end{abstract}

\section{Data-Centric AI Crisis}

Many real objects and processes that humans deal with (human body, biosphere, autonomous transport, social systems, economics etc.) have very high complexity. The complexity is such that human intellectual abilities are not enough to build models of such objects. 
      
At the beginning of this century it became possible to collect and store a large amount of data. By analyzing data about an object collected over a relatively long period of time, it is possible to identify some regularities of behavior, and build its 'behavioral model'. Methods based on data analysis have become the basis for modern data-centric artificial intelligence. The current mainstream of data-centric artificial intelligence methods is machine learning based on neural networks, in particular deep learning. The general scheme of creation of AI based on neural networks is as follows: a representative dataset is collected, which is marked by automatic methods or with human intervention (marked and described the important features in the data), in other words, a dataset is created; a neural network is trained on this dataset, it identifies and 'remembers' the relationship between the features in the data; trained network can use new data of the same structure about the same object to predict the corresponding object characteristics or to model its behavior.

Artificial intelligence based on data is not intelligence in the common sense. Wang~\cite{wang2019} raised a heated debate about redefining artificial intelligence. He proposed to approach intelligence as the ability to adapt when knowledge and resources are lacking. It is important to note that his attempt to define artificial intelligence does not distinguish artificial nature of intelligence. In fact, Wang defines intelligence in general. In this article, following the cognitive sciences, intelligence is defined constructively as a functioning system of cognitive functions. The action of this system allows to extract new knowledge and build new models, which provide the ability to adapt in the lacking of knowledge and resources. Examples of the cognitive functions are perception, attention, memory, language, or planning~\cite{col2010}. Obviously, one of the most important cognitive abilities is abstract thinking. It is based on the ability to replace real world objects and processes with symbols and to operate with these symbols instead of operating with the reality. Nowadays, automatic construction of interpretable symbolic models is impossible. So far, there are no successful models that allow to introduce symbols and give them meaning. Considering the above, it is possible to say that data-centric narrow intelligence is not an intelligence, but rather can be considered as one of the cognitive functions (depending on the subject area, it can be search, classification, translation, etc.). In other words, data-centric intelligence is an advanced stochastic machine. With all benifits, it has a number of significant drawbacks.
\begin{itemize}
        \item  To train a neural network describing a complex object is not always possible to collect and label a sufficient amount of data;
        \item  Training a neural network takes a lot of time and large computational resources;
        \item  For small changes of input data (especially, data structures) a full retraining cycle is required;
        \item  The results produced by a neural network can almost never be interpreted. It is possible to get a result, but have no explanation why this result is correct and what the way it was got. Neural networks always give not an exact, but a probabilistic answer, without an explanatory component they cannot be used for example in fields such as medicine;
        \item  the narrow field of application of solutions based on artificial neural networks. They are not able to solve simple cognitive problems (unable to demonstrate integral cognitive effect), to transfer the results of training to another domain. 
\end{itemize}

Instead of data-centric AI we focus on a class of intelligent systems that allow the seamless integration of human and machine intelligence. This type of intelligence is able to co-evolutionary cognitive development of human and machine agents. The following section describes the main features such systems.

\section{Co-Evolutionary Hybrid Intelligence}

The idea of hybridization of human and machine intelligence is not new. The most significant influence on the development of views on hybridization of in\-telligence was made by Engelbart~\cite{eng1962}. In his framework for augmented intelligence, he defined the capabilities and basic interfaces for human-machine interaction in cognitive tasks. 

Subsequently, Z. Akata and others~\cite{akata2020} defined the concept of hybrid intelligence as a combination of human and machine intelligence, complementing each other. This type of symbiosis makes it possible to achieve goals that are unattainable for either humans or machines separately. 

The mentioned works do not imply general principles of technologization of hybrid intelligent systems creation. We can conclude that, according to the mentiones authors, a hybrid system is created for a specific task. Hence, the way of integration is also chosen in ad-hoc way.

A definition of co-evolutionary hybrid intelligence is given in~\cite{kri2021}. The authors point out the insufficiency of human-machine hybridization at the level of ope\-rations, data and ontologies. The key mechanism for moving towards strong intelligent systems is defined as the ability to build cognitive capabilities in the process of co-evolutionary development. Thus, co-evolutionary hybrid intelligence (CHI) is a symbiosis of artificial and natural intelligence mutually evolving, learning and complementing each other in a process of co-evolution. In this case, co-evolution refers to the ability of the system to change as it functions, based on the knowledge extracted from the domain. 

The main possibility for this kind of co-evolution between man and machine is compatibility at the level of cognitive functions. In other words, if different cognitive abilities can be realized by humans and machines, then it is possible to create an interface that ensures their interchangeability (interoperability). From a pragmatic point of view, functions such as searching, classifying, identifying features in data, translating, and others can be performed by humans and machines. De\-pen\-ding on the amount of data, the level of formalization of the problem, and other aspects, a human or machine will be more efficient. If it is possible to replace one type of agent (human) with another type of agent (machine) and vice versa, it is possible to build hybrid systems. 

A peculiarity of humans is that their performance strongly depends on the degree of fatigue caused by cognitive overload. Also, individual abilities to perform cognitive work matter. Thus, if in a hybrid system a human handles a certain operation noticeably better than a machine, then after a long period of time the performance may become worse than the machine version of the implementation. This fact entails the need to monitor human cognitive abilities during hybrid system operation.

In a broader sense, several classes of interoperability can be considered. Interoperability between the developer and the intelligent system being created defines the possibility to work in a single knowledge ontology of the developer (programmer, data engineer, etc.) with a subject matter expert (e.g. doctor or social engineer). Interoperability between machine and human intelligent agents within a hybrid system provides the ability to jointly extract new knowledge (e.g., software data analysis methods are capable of finding statistically significant patterns, but are not able to interpret them; on the other hand, humans are capable of interpretation, introducing a new symbol into the ontology, but are weak in data analysis). Interoperability between intelligent systems provides the possibility of cooperation between systems created independently (this situation is expected in the near future in the creation of smart cities, saturated with independent devices and services, which are forced to work in a common en\-vi\-ron\-ment).

Technologization of the development of hybrid co-evolutionary intelligent systems relies on the following:
\begin{itemize}
     \item  formalization of cognitive functions that allow the assembly of 'intelligence' implemented by agents of different nature;
     \item  enhancement of capabilities of human-machine interfaces, for knowledge transfer from human to machine and vice versa;
     \item  creation of an individualized model of a human being as a part of a hybrid system, control of his/her state and cognitive abilities (if necessary, reassignment of tasks to other agents)
     \item  minimizing the gap between the "user" and the "developer" of the hybrid system.
\end{itemize}

In addition to the above points, the development of biofeedback methods is important. These methods are well proven in medical applications~\cite{goessl2017, dessy2018, sutaro2013}. Their application allows to use non-verbal mechanisms of self-regulation and state control of human agents.

\section{Cognitive Architectures: State of the Art}

The main proponent of the idea of formulating a theory that would cover all aspects of cognition was Allen Newell, who identified the means to achieve this goal: cognition architectures~\cite{new1990}. The first ideas for creating such architectures can be traced back to Turing's article on the intelligent computer~\cite{tur1950}. Turing believed that speed and memory capacity were the main barriers to the achievement of machine intelligence by computers of the time. History has shown, however, that each advance in artificial intelligence has only clarified how much of the mystery of human intelligence, creativity, and ingenuity is a difficulty~\cite{taa2009}.

There are different approaches for defining "cognitive architecture". Thus, in~\cite{sow2011} the authors suggest that a cognitive architecture is a design of a computing system for modeling some aspects of human cognition. The authors of~\cite{taa2008} believe that cognitive architectures are, on the one hand, part of the original purpose of creating an intelligent machine that exactly replicates human intelligence and, on the other hand, attempts at theoretical unification in the field of cognitive psychology. According to the authors~\cite{lan2009}, expert systems and cognitive architectures are close, but cognitive architectures offer a description of an agent's intellectual behavior at the level of systems, rather than at the level of component methods developed for specialized tasks.

In~\cite{lan2009} there is a more detailed notion of cognitive architecture. Cognitive architecture, as a basic part of an intelligent system, according to the authors, includes those aspects of a cognitive agent that remain unchanged over time and in different fields of application. These typically include: short- and long-term memory, in which the agent's purposes and knowledge are stored; the representation of the elements contained in memory and their larger mental structures; and the processes that operate on these structures, including the learning mechanisms that modify them. The main properties of the cognitive architecture are: knowledge representation, knowledge organization, knowledge utilization, knowledge acquisition and knowledge dissemination.

The cognitive architecture for an intelligent robotic system according to the authors~\cite{bur2005} must support fast perception, control and execution of tasks at a low level, as well as recognition and interpretation of complex contexts, internal task planning and behavioral learning, which are usually handled at higher levels.

Some researchers focus on reinforcing individual properties of AI systems. For example, the authors~\cite{kri2012} argue that autonomy is a key property for any system that can be considered general intelligence. However, today there is no system that combines a wide range of capabilities or presents a general solution for the autonomous acquisition of a large set of skills. The reasons for this are the limited machine learning and adaptation techniques available, and the intrinsic complexity of integrating multiple cognitive and learning capabilities into a whole architecture. The authors consider cognitive architectures in terms of effective implementation of the autonomous properties.

The authors~\cite{how2012} emphasize the importance of the situational awareness module in a cognitive architecture, which includes a broad set of information and analytical requirements. To use it properly, the system must be able to determine the appropriate level of focus for information input at the global, or system level, as well as at the local level, integrating them into a unified picture of the situation. This requires both goal-driven processing and data-driven processing. The former aims at examining the environment according to current unresolved goals, while the latter receives signals from the environment and decides whether new active goals are needed to properly align with the intentions. Dynamic switching between the two models of information processing, according to the authors, is important for successful performance in many environments.

In~\cite{che2018}  the authors present their vision of the approach to creating a hybrid intelligent information system for the basis of cognitive architecture. The authors understand "hybrid intelligent systems" as systems with hybridization of different methods of soft computing, expert systems, neuro-fuzzy systems, fuzzy expert systems, using evolutionary methods to build neural networks and other methods. In their paper the authors consider a simple example of a perception-action cycle for a cognitive architecture based on geo-information system. 

A comparative analysis of the cognitive architectures of Cyc, Soar, Society of Mind and Neurocognitive Networks is given in~\cite{sow2011}, ACT-R, Epic, Soar in~\cite{taa2009}, ACT-R, ICARUS, PRODIGY in~\cite{lan2009}, ACT-R, Soar, LIDA, SiMA, NEF (SPAUN), iCub, SEMLINCS, Summary in~\cite{jim2021}; ACT-R, Soar, NARS, OSCAR, AKIRA, CLARION, LIDA and Ikon Flux in~\cite{kri2012}; Soar, ACT-R, EPIC, Clarion in~\cite{taa2008}, SOAR, ACT-R, CLARION and Vector-LIDA in~\cite{ant2018}.

Analysis of the above literary sources showed that the concept of "cognitive architecture" has existed since the middle of the $20^{th}$ century. This term, in spite of the difference in architecture components, until the current moment is understood as follows: it is a system, which has, to a greater or lesser degree, analogues cognitive functions of a human being. Very rarely is it explicitly said that this architecture interacts with humans. In the rest, just like artificial intelligence, the cognitive system exists by itself. 

The main and very strong distinguish between cognitive architecture presented in this paper and existing ones is the seamless human integration, aimed at the joint development of humans and AI.

\section{Co-Evolutionary Hybrid Intelligence Cognitive Architecture}

\begin{center}
\begin{figure}
\includegraphics[width=11cm]{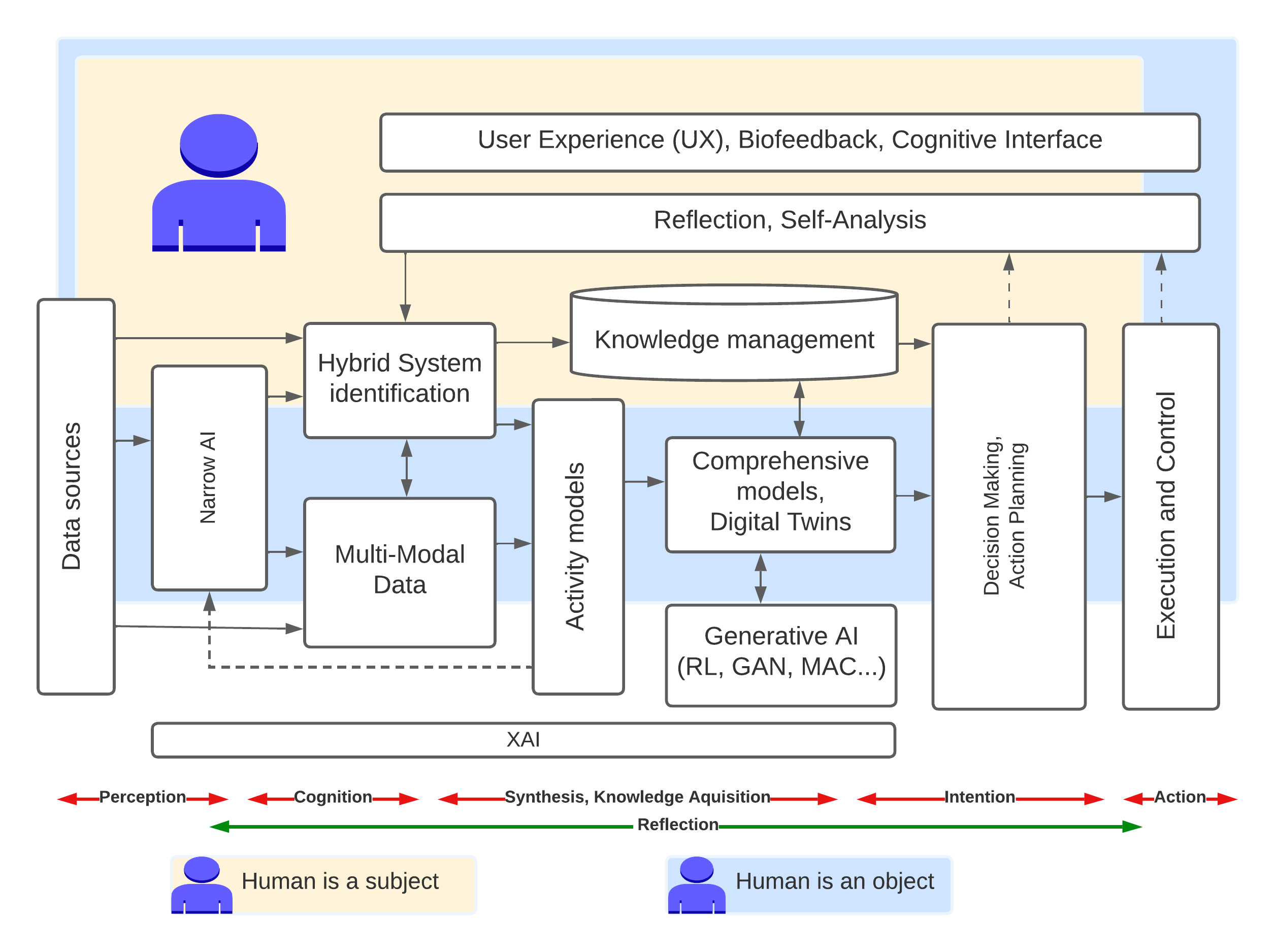}
\caption{Cognitive Architecture for Co-Evolutionary Hybrid Intelligence} \label{arch}
\end{figure}
\end{center}

This section considers cognitive architecture for Co-evolutionary Hybrid Intelligence from a software developer's perspective. Unlike many software frameworks, this architecture must seamlessly include humans (Fig.\ref{arch}). Humans are viewed as both a subject and an object. As a subject, the human acts as an actor in the system and influences how the system works. As a subject, the human is seen as a component of a system with its own dynamic characteristics, which change in the process of operation. In particular, there are periodic states of decreased performance, increased errors due to fatigue and stress. Medically speaking -- humans experience cognitive deficits~\cite{del2014} that affect the overall cognitive power of the system.

Traditionally, we can see a conventional division into different stages of transformation of external signals into recognition of the situation, planning and implementation of actions: perception, cognition, knowledge acquisition and model synthesis, intention and action. All this pipeline also has an obligatory process of self-assessment (or reflection).  Mathematical methods for constructing reflective processes are described in detail by V. Lefebvre~\cite{lef2015}
The functions of the main blocks of the architecture are listed below.
\begin{itemize}
\item \emph{Data Sources}. Various primary data sources (sensors) which receive information about the control object and about the parameters of a human being who is part of the hybrid intelligent system.
\item \emph{Narrow AI}. Data processing techniques including signal cleaning, initial pattern recognition, classification, and approximation. This block contains simple models of observable signals.
\item \emph{Multi-Modal Data}. A generalized data model, containing signals from independent sources, reduced to a single time (and space, where it makes sense). 
\item \emph{Hybrid System Identification}. One of the key building blocks defining a model of a hybrid system as such. To optimize an intelligent system, having its model is required. This block is responsible for accumulation of data and prediction of the hybrid system 'behavior'.
\item \emph{Activity Models}. A person working professionally in some subject area uses two types of knowledge and skills: verbalizable and non-verbalizable. Verbalizable knowledge is comprehended by a person and can be recorded as data or rules. Non-verbalizable knowledge and nonverbalizable experience can only be extracted by observing a person's actions over a long period of time while performing the same procedure (for example: a professional golfer cannot describe the pattern of hitting with a club; a professional radiation diagnostician cannot accurately describe the sequence of processing X-rays to test a hypothesis of a diagnosis). This module is designed to observe the pattern of human actions included in the decision-making cycle in order to extract tacit knowledge. This knowledge obtained from a large number of professionals solving a similar problem can become the basis for the construction of a training system for newcomers in the subject area.
\item \emph{Digital Twins}. Full information about the hybrid system's operation during its lifetime and a set of methods for automatically identifying trends and predicting states.
\item \emph{Generative AI}. Algorithms and methods for generating decisions and directional search for options on a set of hybrid system parameters.
\item \emph{Decision Making and Action Planning}. Scenario planning for management in the short-term and long-term planning horizons.
\item \emph{Execution and Control}. Execution of action scenarios.
\end{itemize}

Separately, it is worth noting the presence of human inclusion components in the hybrid system. These components are aimed at creating the possibility of seamless interaction between machine and human at different levels: neural interfaces, biofeedback, augmented reality, interaction at the level of cognitive functions. Also, the human is part of the reflexive loop of awareness of the possibilities of the hybrid system in which he is a part. 

Treating humans as an integral part of the system has good reasons. The complexity of technical processes in manufacturing is increasing at an exponential rate. On the one hand, this provides humans more opportunities to achieve various goals, to improve their standard of living, to be freed from physically demanding operations in production, to be freed from routine operations or operations requiring computing and other powers beyond the control of the human due to physiological limitations of the human body. On the other hand, the increasingly complex automation of production processes, robotics, and computerization lead to an increasing load on the human cognitive sphere. In other words, all modern technologies, reducing physical load on human organism, lead to increase of psycho-physiological load on human being, forcing him to work at the limit of his psycho-physiological possibilities, in extreme situations.

It is not enough for an intelligent system to be only a tool. To avoid severe consequences from incorrect actions, the development of human cognitive capabilities and artificial intelligence systems cannot be independent of each other. Their coordinated development is required. Such development requires reciprocal feedback between artificial and human manifestations of intelligence. Mutual evaluation of agents in analyzing the situation and choosing the best action is necessary to achieve the goals.

The main characteristics of Co-Evolutionary Intelligence systems as a single working organism are:
\begin{itemize}
    \item mutual learning, when artificial and human intelligent agents complement each other in those areas where their cognitive capabilities are limited due to various reasons;
    \item personification, i.e. tuning of artificial intelligence systems for a definite person (or persons if we are dealing with system with group of human agents), with whom they form a single organism, a system, for achieving certain goals;
    \item condition monitoring of intellectual agents.
\end{itemize}

For the implementation of the last two items it is necessary to introduce modules on monitoring and estimation of a psycho-physiological state of a human into the cognitive architecture.

Following Bartlett's classification\cite{bar1953, bar1951}, it is possible to distinguish two main groups of indicators of the functional states of a person in the loop of production: physiological and psychological. As possible indicators of the dynamics of physical state are considered a variety of types of bio-electrical indicators: EEC, ECG, temperature, pressure and others. 
There is a huge number of devices that allow to measure these indicators. The group of indicators of psychological state dynamics includes performance criteria of different psychometric tests and analysis of subjective symptomatic of specific types of functional states. In most cases during production processes there is no possibility of invasive measurement of physical indicators or passing tests to evaluate psychological criteria.

\section{Conclusion}

Throughout the history of artificial intelligence, it has often been viewed largely as an independent tool. This tool can be trained (or can itself be trained) to perform certain tasks autonomously. From the position of cognitive sciences, intelligence is a system of cognitive functions, capable, among other things, of creating symbolic models and operations with them. Creating symbolic systems for abstract reasoning is the basis for transferring experience from one domain to another and for generalizations. Currently, there are no examples of software or mathematical systems capable of creating meaningful symbolic systems without human intervention. At the same time, in many narrow applications artificial methods are much more efficient than humans in some domains. At development of intellectual systems it makes sense to consider a human as a carrier of some unique cognitive abilities which being integrated with machine methods possess intellectual power surpassing a human and a machine separately. The considered cognitive architecture for co-evolving hybrid intelligence allows to take into account the peculiarities of a human as an intelligent agent. Hybrid evolving systems can become the basis for intelligence superior to human capabilities.

It is also worth to mention, we have a lot of knowledge that we can't impose on intelligent systems. Therefore, mutual learning between humans and machines is a very delicate topic.  It is obvious that the philosophical and ethical exploration can not be ignored, those topics should be one of the core future discussion. A significant issue for ethical research would be the potential for the human-machine interaction process to get out of control and become unsupervised, where the role and importance of intelligent system in the bilateral overall process may become asymmetrically reinforced and acquire the properties of a «defining dominant». In other words, by outsourcing to a machine functions that in the mathematical and algorithmic dimension humans are incapable of performing, there is a risk of gradual loss of control not only over a particular system decision, but also of control over strategic goals, the system of checkpoints ("taboos") and the definition of basic movement coordinates that have always been in human hands when interacting with machinery~\cite{kri2021}.

%
%
%
\bibliographystyle{splncs04}
\bibliography{mybibliography}

\end{document}